# Modeling and parametric optimization of 3D tendon-sheath actuator system for upper limb soft exosuit

Amit Yadav[1,a]; Nitesh Kumar[2,a]; Shaurya Surana[3,a]; Aravind Ramasamy[4,a]; Abhishek Rudra Pal[5,b]; Sushma Santapuri[6,a]; Lalan Kumar[7,c]; Suriya Prakash Muthukrishnan[8,d]; Shubhendu Bhasin[9,c]; Sitikantha Roy[10,a,e,*]

[1] amityadav.9997@gmail.com, [2] nitesh@alumni.iitm.ac.in, [3] shaurya.wpu@gmail.com, [4] 50garavingsamy@gmail.com, [5] abhishekrudrapal@gmail.com, [6] ssantapuri@am.iitd.ac.in, [7] lkumar@ee.iitd.ac.in, [8] dr.suriyaprakash@gmail.com, [9] shubhendu.bhasin@ee.iitd.ac.in, [10] sroy@am.iitd.ac.in

[a] Department of Applied Mechanics, Indian Institute of Technology Delhi, New Delhi, 110016, India
[b] School of Mechanical Engineering, Vellor Institute of Technology Chennai, Tamil Nadu, 600127, India
[c] Department of Electrical Engineering, Indian Institute of Technology Delhi, New Delhi, 110016, India
[d] Department of Physiology, All India Institute of Medical Sciences, New Delhi, 110029, India
[e] School of Artificial Intelligence, Indian Institute of Technology Delhi, New Delhi, 110016, India

**Abstract**

This paper presents an analysis of parametric characterization of a motor driven tendon-sheath actuator system for use in upper limb augmentation for applications such as rehabilitation, therapy, and industrial automation. The double tendon sheath system, which uses two sets of cables (agonist and antagonist side) guided through a sheath, is considered to produce smooth and natural-looking movements of the arm. The exoskeleton is equipped with a single motor capable of controlling both the flexion and extension motions. One of the key challenges in the implementation of a double tendon sheath system is the possibility of slack in the tendon, which can impact the overall performance of the system. To address this issue, a robust mathematical model is developed and a comprehensive parametric study is carried out to determine the most effective strategies for overcoming the problem of slack and improving the transmission. The study suggests that incorporating a series spring into the system's tendon leads to a universally applicable design, eliminating the need for individual customization. The results also show that the slack in the tendon can be effectively controlled by changing the pretension, spring constant, and size and geometry of spool mounted on the axle of motor.

---

[*] Corresponding Author

## 1. Introduction

Soft wearable robots have the potential to enhance the capabilities of the human body through the use of flexible, lightweight and advanced actuation systems. The appearance of the overall system is largely dependent on the choice of actuators and transmission mechanisms. To reduce the inertia and weight of the wearable device, it is often necessary to locate the actuators in a remote location. There are options for actuation systems in soft wearable robots, such as fluidic actuators [1–3] and tendon-driven actuators [4,5]. Tendon-driven actuation is a promising approach for developing soft wearable robots, as it allows for the use of small, lightweight motors and enables the robot to adapt to the movement of the wearer. A single motor can be used to actuate the upper limb, provided that a proper understanding of the mechanics underlying the system is achieved.

Several previous studies have addressed the mechanical behavior of soft wearable robots in few years. Phee *et al.* [6] presented a mathematical model to estimate the force at distal end of a tendon sheath actuator and validated the results with experiments. Hellman and Santos [7] introduced a unique concept for a motor-based actuation system for tendon-driven robotic hands with N-type or 2N-type control setups. The system was constructed as a prototype and its dynamic response was the subject of investigation, including the effect of different spring constants on its natural frequency. Lin and Wang [8] presented a study of the transmission characteristics of a single-tendon sheath actuation system, including the effects of friction and tendon compliance on performance.

Wang *et al.* [9] studied the friction and tension transmission characteristics of flexible and time-varying tendon-sheath configurations and developed models to accurately predict the transmission rates in these challenging, yet common real-world applications. The proposed models can be used to design and control remote mechanical actuation systems in robotic devices, allowing for efficient energy transmission through narrow working channels and taking into account the length effect on the transmission rate. Sun *et al.* [10] developed a model for accurately estimating the position of a robotic system using a flexible tendon-sheath actuation mechanism, without the need for sensors at the distal end. The research lay the groundwork for any robotic system employing a tendon-sheath actuation mechanism to achieve precise position control.

Do *et al.* [11] introduced a novel dynamic friction model for tendon-sheath actuated surgical systems. The model successfully captured nonlinear hysteresis and accurately predicted tension and friction forces at different displacements. Lau *et al.* [12] developed a motion compensation model for a cable driven continuum manipulator and found to reduce position error by less than 5%. The model can be applied in an endoscopic surgical robot to evaluate its effectiveness in improving the accuracy and completion time of surgical tasks. A novel tension and displacement transmission model for a cable-pulley system in laparoscopic surgical robots was presented by Xue *et al.* [13]. Experimental validation confirmed the accuracy of the model and demonstrated improved performance through a real-time compensation control algorithm, effectively reducing backlash angles and tracking position errors.

Wagner and Emmanouil [14] presented the pros and cons of using electrical wire transmissions and tendon-sheath mechanical transmissions in surgical robotics, considering factors such as efficiency, force, power, material properties, access geometry, and safety limits. The results showed that electrical transmissions can effectively deliver the required power for surgery, which could lead to the creation of more minimally invasive surgical devices. Shao *et al.* [15] presented a new artificial tendon actuator inspired by the Hill-type muscle model, which had variable compliance properties using two linear springs. For displacement and force control without distal feedback, a model-based inverse control method with bending angle compensation was developed, and the efficacy of the suggested control technique was proved by trajectory tracking tests. A two-degree-of-freedom soft exoskeleton that can help with simultaneous elbow and wrist mobility was developed by Ismail *et al.* [16]. The device may offer mechanical help for pronation and supination angles up to 19° and 18°, respectively, as well as flexion and extension motions ranging from 90° to 157°.

Jia *et al*. [17] proposed a new dual tendon sheath transmission technique that provides long-distance and large-stroke power transfer with increased precision for robots functioning in nuclear magnetic environments. The experimental results were used to build and validate the analytical friction and hysteresis model for the system. Yin et al. [18] developed a new method for compliant control in robotics using tendon-sheath actuators that cannot accommodate force sensors due to space limitations. A soft artificial muscle that is very compliant, simple to fabricate, and operate similarly to human muscle was developed by Phan *et al.* [19]. It was concluded that the future research should concentrate on developing new silicone materials with higher stress and strain, as well as using sophisticated nonlinear hysteresis models to capture high strain behavior.

Ennaiem *et al.* [20] examined the repeatability of upper limb motion during specific rehabilitation exercise, analyzing five participants' movement patterns. The findings inform that the optimal design of a cable-driven parallel robot for upper limb rehabilitation, considering workspaces identification and structural optimization. Lee *et al*. [21] proposed a simpler piecewise linear model for building backlash hysteresis as well as practical approach for identifying model parameters by employing motor current from a controller. This model demonstrated a considerable decrease in errors and enhanced robotic control accuracy with this new approach. A real time position compensation control approach for manipulators equipped with a dual tendon-sheath system was developed by Wu *et al*. [22]. The results demonstrated that the developed control approach could achieve the desire torque trajectory even when subjected to external disturbances. Jung and Bae [23] proposed a series elastic tendon-sheath actuation mechanism for multi-degrees of freedom systems. The mechanism featured series elastic elements and a feedforward controller to enable accurate torque control.

A novel class of fluidic soft, scalable soft actuator that can be integrated into a wearable fabric sleeve for the upper limb augmentation was introduced by Hoang et al [24]. Kernot and Ulrich [25] suggested to utilize a cable-driven manipulator to carry out non-cooperative capture and it was found that using adaptive control techniques improves the closed-loop gripper's tracking performance. In a systematic review of soft robotic devices for upper limb assistance, Bardi *et al*. [26] discovered that the pneumatic and cable driven actuators are most often used auction system. A general compensating control approach for flexible manipulators operated by a cable-driven mechanism was proposed by Zhang [27]. The technique addresses the nonlinear and hysteretic properties for enhanced motion accuracy and may be used with any flexible manipulators. In order to provide precise and reliable force control during actual human-robot contact, Han et al. [28] presented a model-free force controller for actuator that makes use of an improved extended state observer to obtain the unknown disturbance terms and regulate input gain. Zhang *et al*. [29] developed a mathematical model to describe hysteresis in tendon-sheath mechanisms utilized in nature orifice transluminal endoscopic surgery and flexible endoscopy. By implementing the proposed feedforward compensator, the tendon sheath system was efficiently linearized and enhanced for the positioning accuracy in robot-assisted surgical procedures. Yuan *et al*. [30] introduced a modular muscle system for musculoskeletal robots that can be deployed remotely and transmit muscle tension along small routes. The end-point three-dimensional trajectory tracking of the arm was enhanced by the feed-forward multi-layer neural network technique, which accounts for friction. Wu *et al.* [31] presented a novel tendon-based mechanism for finger flexion and extension inspired by the human musculoskeletal system. The experimental evaluation on an artificial finger showed that the proposed mechanism successfully prevented joint hyperextension and provided a flexion trajectory closer to voluntary flexion motion. Ramasamy [32] proposed an innovative exosuit design based on tendon sheath actuation and the hill muscle model to better mimic human muscle actuation.

Despite these advances, a more comprehensive study should be carried out to investigate the effect of different design and operational parameters on the transmission behaviour of the tendon sheath actuator utilized for upper limb augmentation. A mathematical model is developed for the double cable-driven actuator, taking into consideration the intrinsic characteristics and friction losses. An experimental characterization is performed to determine the friction coefficient between the tendon (wire) and tendon-sheath (conduit). The proposed mathematical model not only assists in designing a slack-free system, but it also serves as a guide for the development of control algorithms to regulate the tension output of the system.

## 2. Mathematical formulation

The total number of motors in a soft wearable tendon-driven robot can be reduced if a single motor can drive multiple tendons simultaneously. This is achieved by utilizing the multiple spools in an array, potentially increasing the compactness and efficiency of the wearable soft robot. The arrangement of the proposed tendon sheath actuator for upper limb augmentation using a single motor mounted on the back of the wearer is shown in Fig. 1. The spool attached to the motor's axle consists of two segments - one for the extensor cable and the other for the flexor cable. The tendons/cables are wrapped in opposite directions so that when the axle of the motor rotates, one cable pulls the arm from one side while the other cable releases it from the other side. Fig. 3 illustrates the proposed spools with two segments. The segment of spool that carries the extensor and flexor cables are named as extensor spool and flexor spool, respectively, in this paper for simplicity.

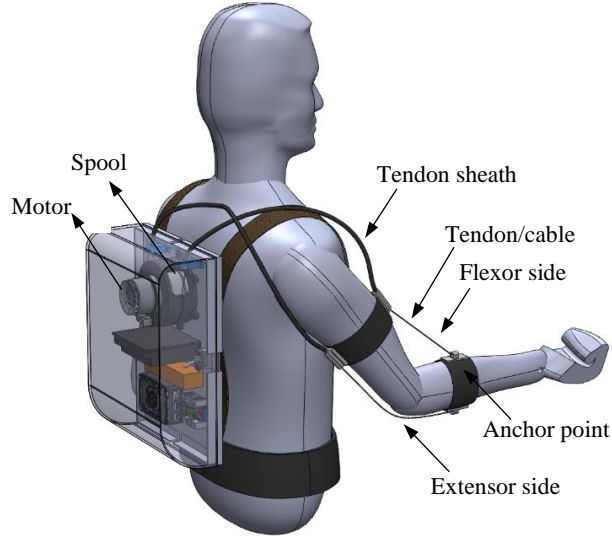

**Fig. 1.** Tendon sheath actuator for upper limb augmentation

*2.1.Two spools with one constant and other with varying radius*

The routing of tendon-sheath around the upper limb and the geometrical parameters related to human arm are considered as shown in Fig. 1 and 2, respectively. The position vectors of points $p$, $q$, $r$ and $s$ are $p = [b_2 \sin\theta \quad -b_2 \cos\theta]$, $q = [b_2 \sin\theta + a_2 \cos\theta \quad -b_2 \cos\theta + a_2 \sin\theta]$, $r = [a_1 \quad b_1]$, and $s = [0 \quad b_1]$. The length between two anchor points $q$ and $r$ on flexor side changes with the change in angle $\theta$ and it can be expressed as,

$$L_c = \left( (b_2 \sin\theta + a_2 \cos\theta - a_1)^2 + (a_2 \sin\theta - b_2 \cos\theta - b_1)^2 \right)^{1/2}. \tag{1}$$

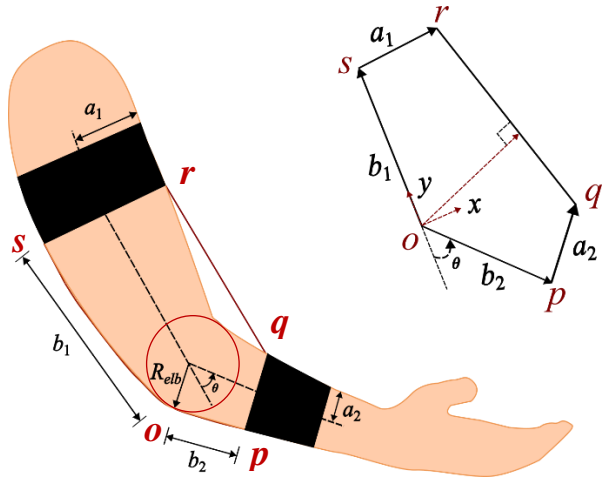

**Fig. 2.** Human arm model with considered geometrical parameters.

The wrapped length of the tendon on spool during flexion is $H_f = (b_1+b_2)-L_c$ and corresponding unwrapped length on extensor side is $H_e = R_{elb}\theta$. Assume that the spool rotates by an angle $\alpha$; the following relationships for two segments of spool (see Fig. 3) can be established:

$$\alpha = \frac{(b_1+b_2)-L_c}{R_m^e} \text{ and } \alpha = \frac{R_{elb}\theta}{R_m^f}, \qquad (2)$$

where, $R_m^e$ and $R_m^f$ are the radii of spools winding or unwinding the extensor and flexor tendon, respectively. The ratio of the two radii to maintain the continuous tightening or no slack condition in the tendon during the elbow movement, can be determined using Eqs. (1) and (2) as,

$$\frac{R_m^e}{R_m^f} = \frac{(b_1+b_2)-\left((b_2\sin\theta+a_2\cos\theta-a_1)^2+(a_2\sin\theta-b_2\cos\theta-b_1)^2\right)^{1/2}}{R_{elb}\theta}. \qquad (3)$$

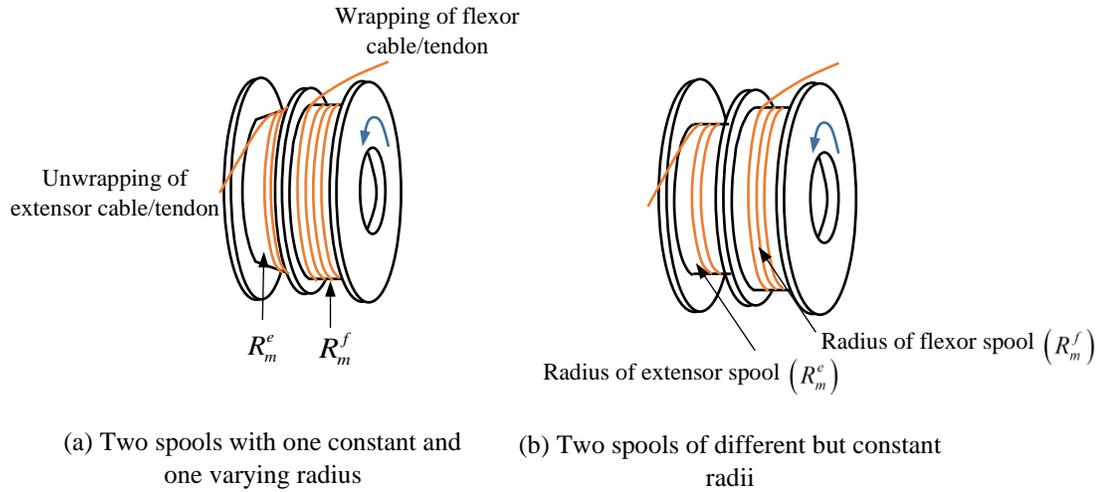

(a) Two spools with one constant and one varying radius

(b) Two spools of different but constant radii

**Fig. 3.** Spools mounted on the axel of motor for flexion and extension of arm

Fig. 4 illustrates the change in the radius of extensor spool as the elbow angle varies, with respect to the radius of flexor spool being kept constant. It shows that the radii ratio for elbow angle, $\theta = \pi/2$ is greater than one, and as the elbow angle decreases, this ratio decreases as well.

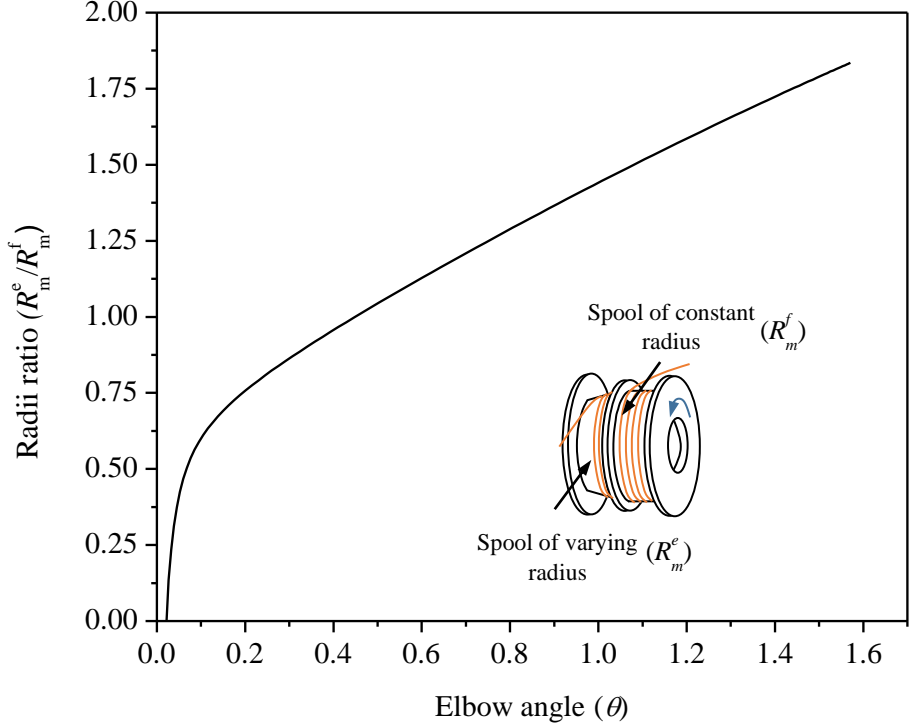

**Fig. 4.** Outer radius ($R_m^e$) to inner radius ($R_m^f$) ratio of spool for no slack condition

The required motor torque can be written as,

$$(F_I^f R_m^f - F_I^e R_m^e) = \tau_M, \qquad (4)$$

where, $F_I^f$ and $F_I^e$ are the tension in flexor and extensor cables, respectively, near motor.

Assistive torque at elbow can be expressed as,

$$\tau_a = J^f(\theta_{elb})F_o^f + J^e(\theta_{elb})F_o^e, \qquad (5)$$

where, the notations $J^f(\theta_{elb})$ and $J^e(\theta_{elb})$ depict moment arm in flexor and extensor side of the arm, respectively. Tension in the tendon on flexion and extension sides can be expressed as,

$$F_o^f = F_I^f \alpha_t \quad \text{and} \quad F_o^e = F_I^e \alpha_r, \qquad (6)$$

where $\alpha_t = e^{-\mu \int_0^L k(s).ds} = e^{-\mu \phi_{bend}^{(f)}}$ and $\alpha_r = e^{\mu \int_0^L k(s).ds} = e^{\mu \phi_{bend}^{(e)}}$. A detailed mathematical derivation of expressions given in Eq. (6) is given in Appendix A.

Substituting expressions form Eq. (6) into (5) yields,

$$\tau_a = J^f(\theta_{elb})F_I^f \alpha_t + J^e(\theta_{elb})F_I^e \alpha_r. \qquad (7)$$

During the flexion and extension of the arm, the tendon on the flexor side undergoes tension while the tendon on the extensor side experiences no tension. Hence,

$$F_I^e = 0. \tag{8}$$

Substituting above expression in Eq. (7) yields,

$$\tau_a = J^f(\theta_{elb}) F_I^f \alpha_t. \tag{9}$$

Similarly, Eq. (4) can be written as,

$$\tau_m = F_I^f R_m^f. \tag{10}$$

Solving Eq. (9) and (10) together gives following relation between motor torque and assistive torque,

$$\tau_m = \tau_a \frac{R_m^f e^{\mu \phi_{bend}^f}}{J^f(\theta_{elb})}. \tag{11}$$

The torque required at motor can be estimated using Eq. (11) which is useful for development of control algorithms. The calculation indicates that the extensor spool's radius needs to vary with the elbow angle, but this would be a challenging to implement in a physical model, and it would not result in a durable design. Another option would be to utilize two spools with constant radii that are different in radius from each other, and these spools would hold the series-connected spring (see Fig. 5) with the tendons that have been pretensioned.

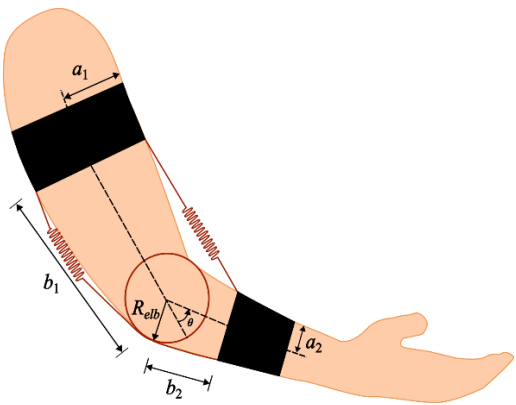

**Fig. 5.** Series-connected spring to tendon sheath system

## 2.2. Two spools of different but constant radii

The concept of utilizing two spools with distinct but constant radii for wrapping and unwrapping of the tendon during limb movement is considered. The ratio of these radii is represented as $\eta = R_m^e / R_m^f$, where, $R_m^e$ and $R_m^f$ are the radii of extensor and flexor spools, respectively. A mathematical relation based on the idea that the length wrapped by flexor spool must equal to the length unwrapped by extensor spool is established as,

$$\eta \left[ \Delta L_{Ten}^f + \Delta L_{SE}^f \right] + \Delta L_{Ten}^e + \Delta L_{SE}^e = \eta H_f + H_e, \tag{12}$$

where, $\Delta L$ and $H$ denotes the stretching and extension function, respectively. The expressions for extension functions are discussed in section 2.1. The subscript '*Ten*' and '*SE*' represent the tendon and series spring, respectively. The sub/superscript *f* and *e* are used for flexor and extensor side of the arm, respectively. The stetching in tendon at flexor and extensor side can be expressed as,

$$\Delta L_{Ten}^f = \frac{F_I^f \beta_t}{AE} - \frac{F_o^f \beta_t}{\alpha_t AE} \quad \text{and} \quad \Delta L_{Ten}^e = -\frac{F_I^e \beta_r}{AE} + \frac{F_o^e \beta_r}{\alpha_r AE}. \tag{13}$$

where, $\beta_t = L \left[ \dfrac{1 - e^{-\mu \phi_{bend}^{(f)}}}{\mu \phi_{bend}^{(f)}} \right]$ and $\beta_r = L \left[ \dfrac{1 - e^{\mu \phi_{bend}^{(e)}}}{\mu \phi_{bend}^{(e)}} \right]$.

The stretching in springs at both sides is taken as,

$$\Delta L_{SE}^f = \frac{F_I^f \alpha_t}{K_{SE}^f} - \frac{F_o^f}{K_{SE}^f} \quad \text{and} \quad \Delta L_{SE}^e = \frac{F_I^e \alpha_r}{K_{SE}^e} - \frac{F_o^e}{K_{SE}^e}. \tag{14}$$

Substitutig the expressions from Eqs. (13) and (14) into Eq. (12) yields:

$$\eta \left[ \frac{F_I^f \beta_t}{AE} - \frac{F_O^f \beta_t}{\alpha_t AE} + \frac{F_I^f \alpha_t}{K_{SE}^f} - \frac{F_O^f}{K_{SE}^f} \right] - \frac{F_I^e \beta_r}{AE} + \frac{F_O^e \beta_r}{\alpha_r AE} + \frac{F_I^e \alpha_r}{K_{SE}^e} - \frac{F_O^e}{K_{SE}^e} = \eta H_f + H_e. \tag{15}$$

Simplifying Eq. (15) results in

$$\eta F_I^f \xi_t + F_I^e \xi_r = \eta H_f + H_e + F_O^e \gamma^e + \eta F_O^f \gamma^f, \tag{16}$$

where, $\xi_t = \left[ \dfrac{\beta_t}{AE} + \dfrac{\alpha_t}{K_{SE}^f} \right]$, $\xi_r = \left[ -\dfrac{\beta_r}{AE} + \dfrac{\alpha_r}{K_{SE}^e} \right]$, $\gamma^f = \left[ \dfrac{1}{K_{SE}^f} + \dfrac{\beta_t}{\alpha_t AE} \right]$ and $\gamma^e = \left[ \dfrac{1}{K_{SE}^e} - \dfrac{\beta_r}{\alpha_r AE} \right]$.

Torque at motor end can be expressed as,

$$\left(F_I^f - \eta F_I^e\right) R_m = \tau_m, \tag{17}$$

where, $F_I^f$ and $F_I^e$ are tension in tendon in flexion and extension side, respectively. Substituting the expression for $F_I^f$ from Eq. (17) into Eq. (16) yields,

$$\eta \xi_t \frac{\tau_m}{R_m} + F_I^e \left[\eta^2 \xi_t + \xi_r\right] = H_{ef} + \lambda_{ef}, \tag{18}$$

where,

$$\lambda_{ef} = F_o^e \gamma^e + \eta F_o^f \gamma^f \text{ and } H_{ef} = \eta H_f + H_e. \tag{19}$$

The expressions for $F_I^e$ and $F_I^f$ are derived from Eqs. (17) and (18) as,

$$F_I^e = \frac{(H_{ef} + \lambda_{ef})}{\left[\eta^2 \xi_t + \xi_r\right]} - \frac{\eta \xi_t \tau_m}{R_m \left[\eta^2 \xi_t + \xi_r\right]} \text{ and } F_I^f = \frac{\tau_m \xi_r}{R_m \left[\eta^2 \xi_t + \xi_r\right]} + \frac{\eta (H_{ef} + \lambda_{ef})}{\left[\eta^2 \xi_t + \xi_r\right]} \tag{20}$$

The tension in tendon near elbow and motor can be determined by using the above expressions.

## 3. Results and discussion

Tendon sheath actuators has various applications, including rehabilitation, therapy, and industrial automation. The study focuses on the double tendon sheath system, which features two sets of cables - one for the agonist and one for the antagonist side - to achieve smooth and natural movements of the arm. The performance of an exo-suit equipped with a single motor is evaluated and characterized through analytical approach to identify any limitations or challenges. The following properties for tendon sheath system are considered in this study: The spring constant of series spring is 3 kN/m; the friction coefficient between tendon and sheath is 0.07. The numerical results are obtained for a tendon routed in a two dimensional plane which is parallel to sagittal plane and touches the arm. The bend angle, length and diameter of the tendon are considered as $\pi$, 2 m and 1.5 mm, respectively. The radius of flexor spool mounted on the motor is 30 mm. The Young's modulus of tendon is $1.45 \times 10^{11}$ N/m² and it is subjected to a pretension of 100 N unless it is mentioned otherwise. The radius of virtual pulley at the elbow joint as shown in Fig. 2, is 45 mm. The values of parameter related to human arm, $a_1$, $b_1$, $a_2$, and $b_2$ are 40 mm, 150 mm, 20 mm and 150 mm, respectively.

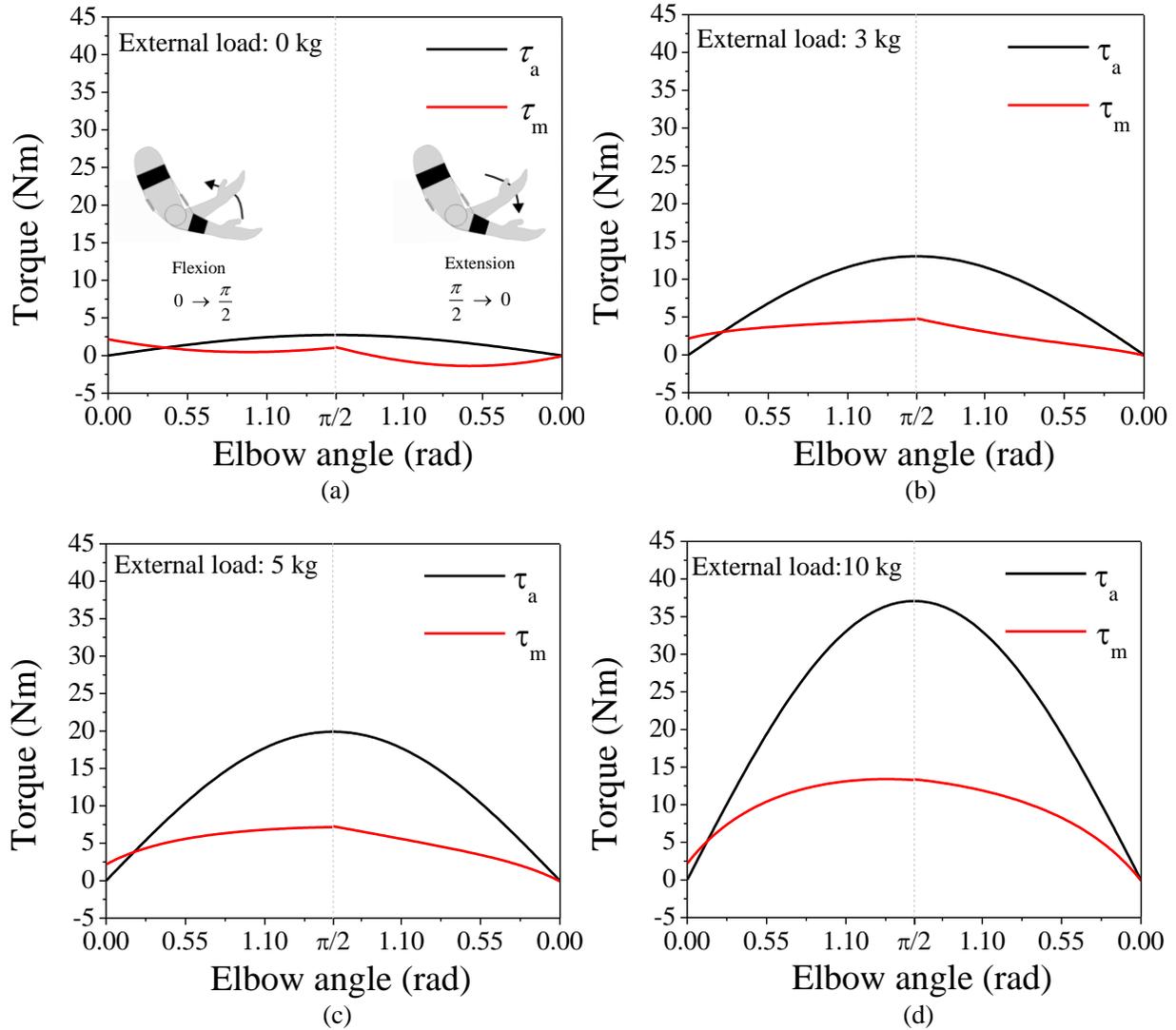

**Fig. 6.** Torque required at elbow joint ($\tau_a$) and motor ($\tau_m$) with respect to elbow angle ($\theta$) for uplifting external loads of (a) 0 kg, (b) 3 kg, (c) 5 kg and (d) 10 kg.

Fig. 6 presents the torque required at the elbow and motor to lift various amounts of external weights (i.e., 0 kg, 3 kg, 5 kg, and 10 kg) as the elbow angle changes during flexion and extension of arm. The tendon sheath system includes a pretensioned tendon and series springs with a spring constant of 30 kN/m, and a single spool with a constant radius of 30 mm. The results depicted in the Fig. 6 reveal that the torque required at the elbow joint consistently exceeds the torque required at the motor, and this difference increases as the weight of the external load increases. However, it is worth noting that the figure shows a few initial values during the flexion movement where this trend deviates. This deviation can be attributed to the pretension applied in the tendon system.

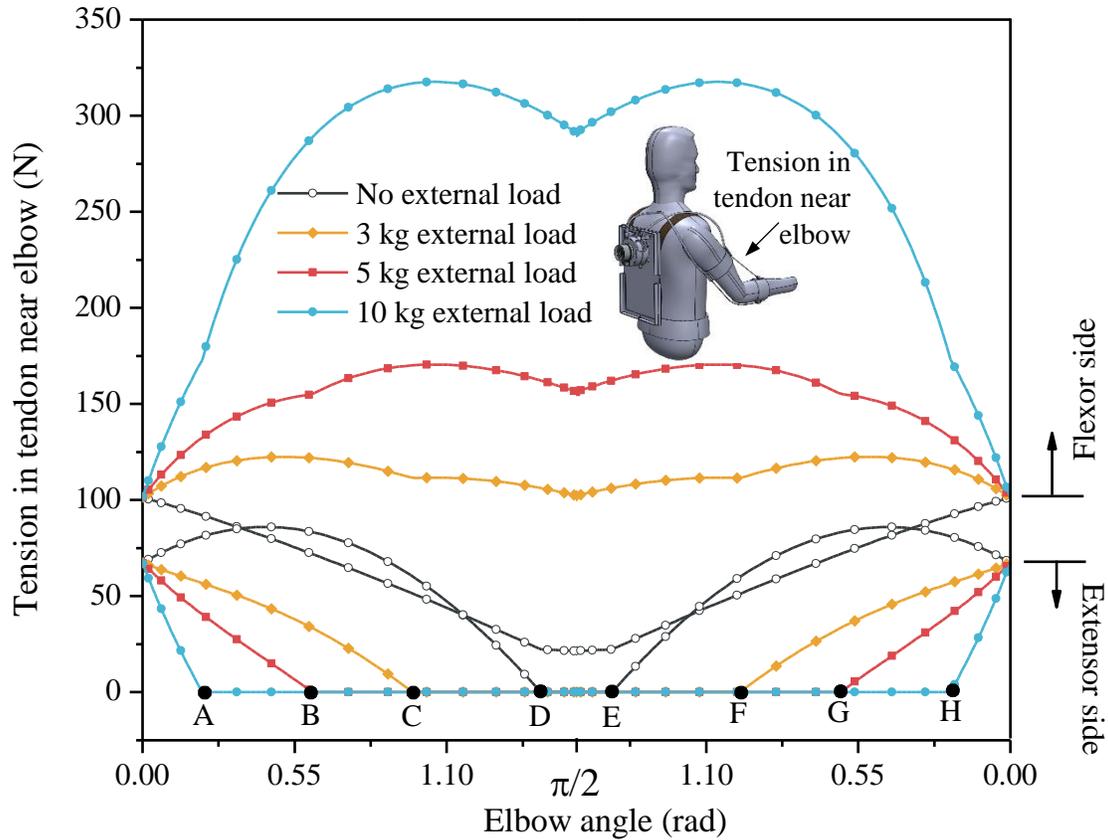

**Fig. 7.** Tension in tendon near elbow at flexion and extension side of the upper limb with respect to elbow angle ($\theta$) while uplifting external loads of 0 kg, 3 kg, 5 kg and 10 kg.

Fig. 7 illustrates the tension in the tendon near the elbow joint at the front and back side of the arm with respect to the elbow angle for various external loads. The results depicted in the Fig. 7 reveal a symmetrical pattern across all cases. It is observed that the tension in the tendon at the front side of the arm consistently decreases as the elbow angle increases during flexion when the arm is not subjected to any external load. However, an initial increase in tension is observed during flexion when the arm is subjected to external loads. Furthermore, the tension in the tendon on the back side of the arm consistently decreases, with the exception of one case (when no external load is applied) where an initial increase is observed. This decrease in tension with the increase in elbow angle results in slack in the tendon on the back side of the hand. It is important to note that the tendon system is not capable of handling compressive forces and thus, such slack should be avoided to ensure optimal functioning of the joint.

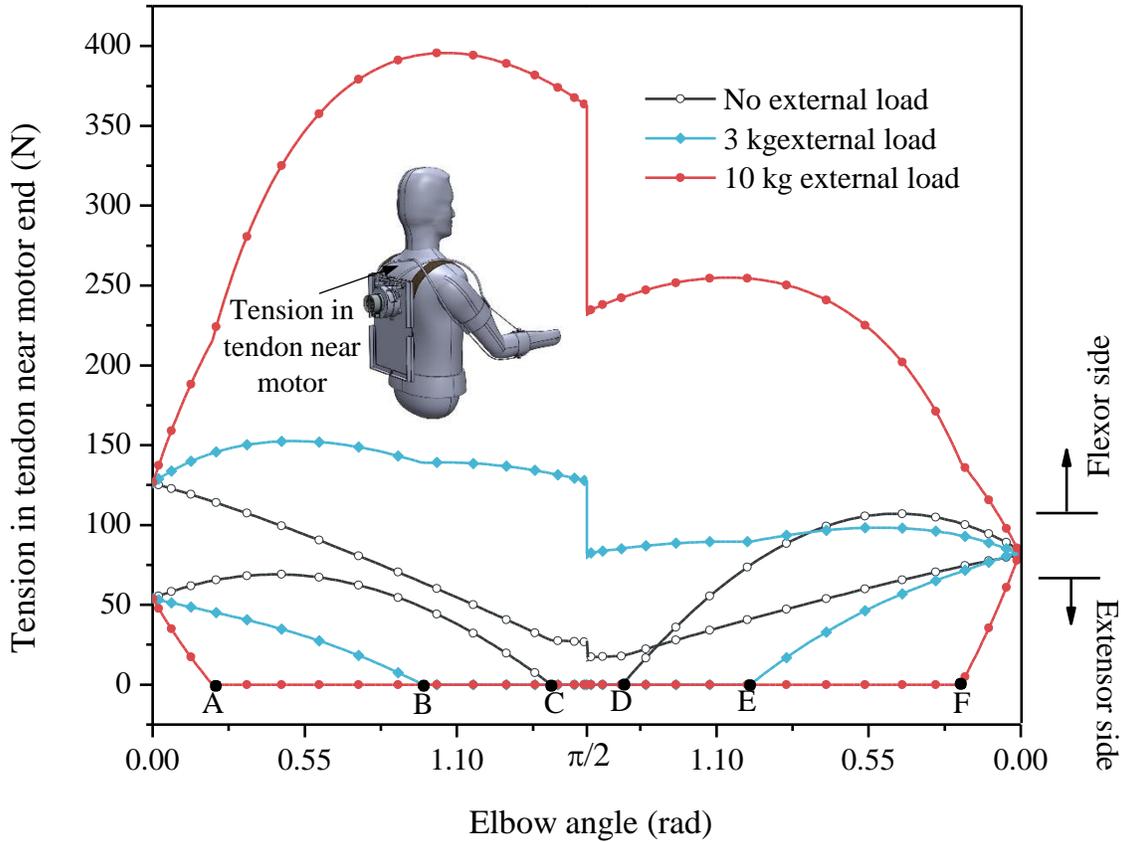

**Fig. 8.** Tension in tendon near motor end at flexion and extension side of the upper limb with respect to elbow angle ($\theta$) while uplifting external loads of 0 kg, 3 kg and 10 kg.

The tension in tendon near motor end versus elbow angle is plotted in Fig. 8 for different external loads. Unlike the results presented in the previous figure, the results in Fig. 8 exhibit an unsymmetrical pattern across all cases. This deviation from symmetry is a result of change in the direction of friction in the tendon sheath system. The tendon routed on the front side of the arm experiences the tension in all cases however, the slack in tendon on the back side of the arm can be seen on motor end.

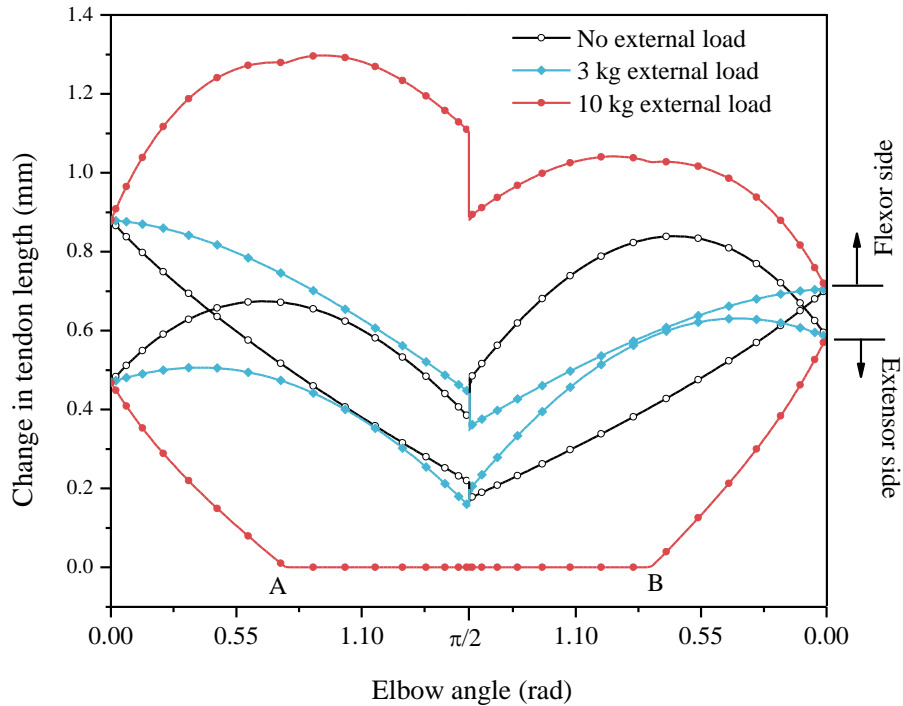

**Fig. 9.** Total change in tendon length at flexion and extension side of the arm with respect to elbow angle ($\theta$) while uplifting external loads of 0 kg, 3 kg and 10 kg. [Pretension = 200N]

In an effort to address the issue of slack in the tendon and spring on the extension side, the value of pretension was increased from 100 N to 200 N. The resulting change in the lengths of the tendon is depicted in Fig. 9. The results indicate that the problem of slack is partially resolved, as it is not observed in two cases (i.e., 0 kg and 3 kg), however, it is still present when the external load amplitude reaches 10 kg. While increasing the value of pretension can further mitigate this issue, it also introduces new challenges. For instance, a person using this system for rehabilitation may not feel the pressure in the forearm, which can lead to impaired blood flow in the arm.

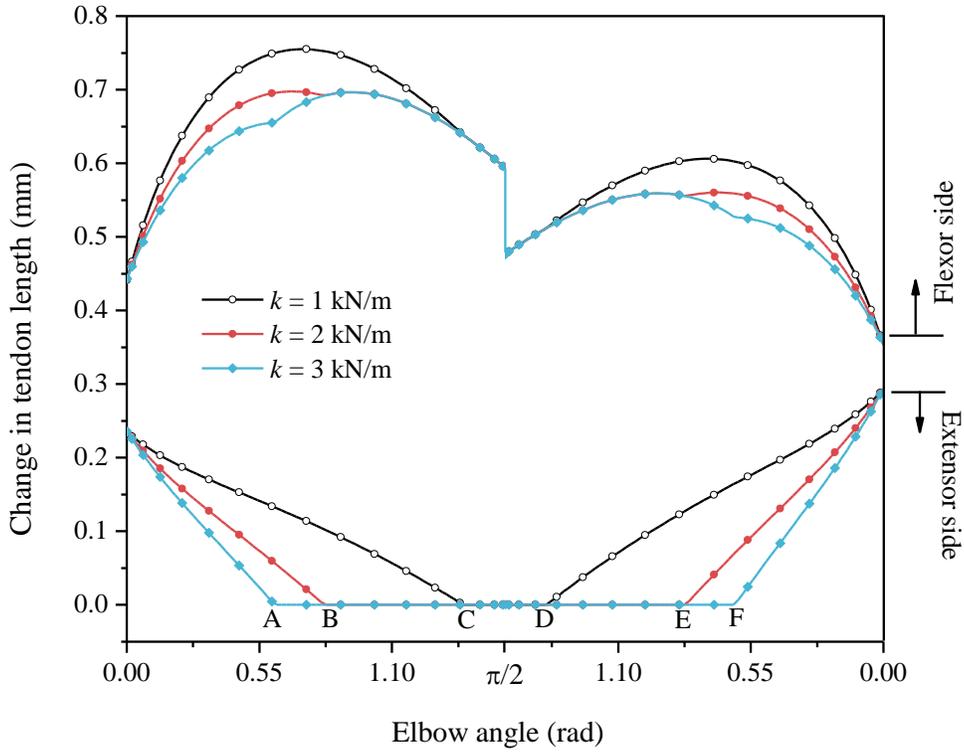

**Fig. 10.** Total change in tendon length for different spring constants at flexion and extension side of the arm with respect to elbow angle ($\theta$) while uplifting 5 kg external load. [Pretension = 100N]

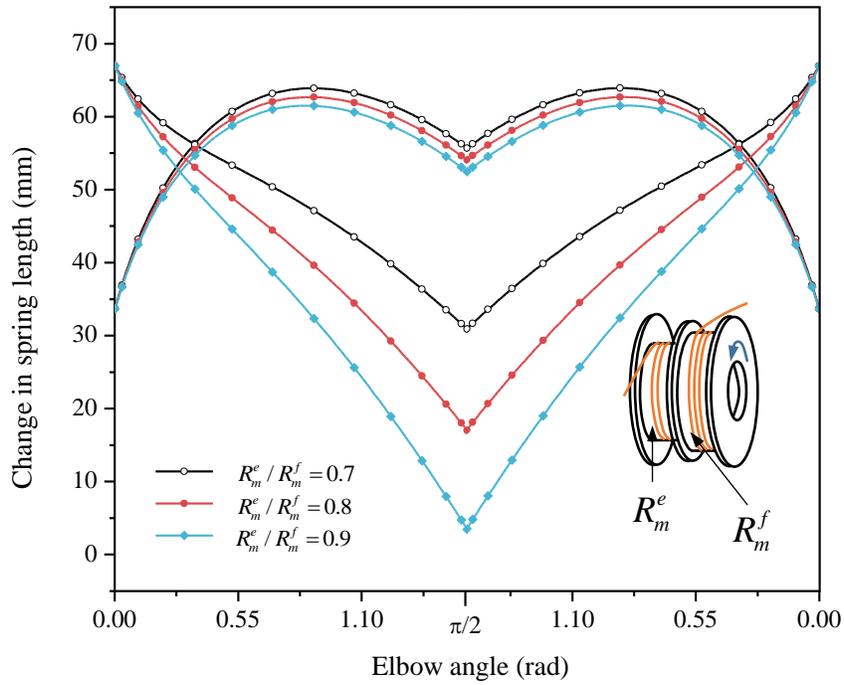

**Fig. 11.** Total change in spring length for different radii ratio at flexion and extension side of the arm with respect to elbow angle ($\theta$) while uplifting 5 kg external load. [Pretension = 100N]

In order to resolve the issue of slack in the tendon and spring system without increasing the pretension, the spring constant was adjusted and the resulting changes in the lengths of the tendon is presented in Fig. 10. The results indicate that as the value of the spring constant decreases, the slack in the system also decreases. However, it should be noted that the spring constant cannot be reduced to a very small value as it may result in excessive extension, which is not permissible within the constraints of the fabricated system.

The slack observed in the tendon and spring system on the back side of the hand is caused by the use of a single spool on the motor that wraps the tendon from the front side during flexion and unwraps the same length on the back side of the hand. In order to reduce the amount of length unwrapped on the back side, the use of two spools with constant but different radii is proposed. Specifically, the spool radius for wrapping should be greater than the radius of the unwrapping spool. Three cases were studied, where the ratio of spool radii were varied at 0.7, 0.8, and 0.9. The resulting changes in the lengths of the spring are presented in Fig. 11. The results indicate that the use of two spools of different radii effectively eliminates the slack in the system.

**Conclusion**

In this study, the torque required at the elbow joint and motor, the tension in the tendon near the elbow joint and motor end, the extension of the tendon and spring, and the impact of pretension, spring constant, and spool radius on the mechanical behavior of the tendon and spring system are investigated. The results obtained from the study reveal several key observations:

1. The torque required at the elbow joint consistently exceeded the torque required at the motor end. This difference increased as the weight of the external load increased.
2. The tension in the tendon near the elbow joint exhibited a symmetrical pattern across all cases, except for some deviation observed during the transition from flexion to extension, likely due to the presence of friction in the tendon system.
3. The extension of the tendon and spring under tensile loads showed a similar pattern of change and the deviation from symmetry is likely due to friction.
4. The problem of slack in the tendon and spring on the extension side was partially resolved by increasing the value of pretension, however, this also introduces new challenges such as impaired blood flow in the arm.

5. The problem of slack was also solved by adjusting the spring constant, but it should be noted that the spring constant cannot be reduced to a very small value as it may result in excessive extension.
6. The slack observed in the tendon and spring system on the back side of the hand was due to a single spool on the motor. It was reduced by using two spools of constant but different radii, in which the spool radius for wrapping was kept greater than the radius of the unwrapping spool.
7. The results of this study demonstrate the importance of considering the mechanical behavior of the tendon and spring system under different loading conditions in order to design effective rehabilitation devices or prosthetic arms.

The study also highlights the trade-offs that must be considered when adjusting the pretension, spring constant, and spool radius in order to reduce slack without introducing other problems. These findings can be used as a reference for future research on the optimization of the tendon and spring systems in rehabilitation devices.

**Appendix A**

**1. A novel three-dimensional mathematical model for distal end tension in a tendon-sheath actuator system**

A small element of tendon sheath having length '*ds*' is considered as shown in Fig. 12.

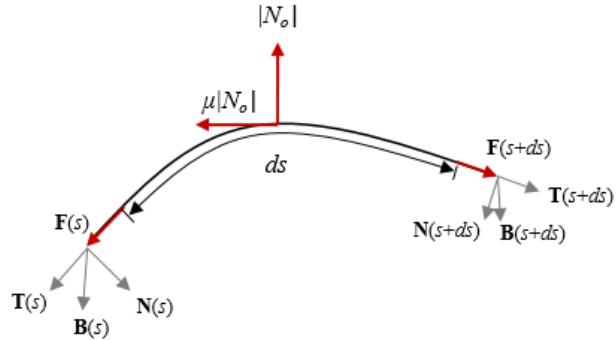

**Fig. 12.** A small element of tendon sheath system

Here, $F(s)$ is the tension in tendon; $T(s)$, $B(s)$ and $N(s)$ denote the Frenet–Serret triad; $\mu$ is coefficient of friction.

$$-F(s)\hat{\mathbf{T}}(s)+|\mathbf{N}_o|\left(\hat{\mathbf{N}}(s)+\frac{d\hat{\mathbf{N}}(s)}{ds}\cdot\frac{ds}{2}\right)+F(s+ds)\hat{\mathbf{T}}(s+ds)-\mu|\mathbf{N}_o|\left(\hat{\mathbf{T}}(s)+\frac{d\hat{\mathbf{T}}(s)}{ds}\cdot\frac{ds}{2}\right)=0 \quad (A.1)$$

By adopting Taylor's series expansion,

$$-F(s)\hat{\mathbf{T}}(s)+|\mathbf{N}_o|\left(\hat{\mathbf{N}}(s)+\frac{d\hat{\mathbf{N}}(s)}{ds}\cdot\frac{ds}{2}\right)+\left(F(s)+\frac{dF(s)}{ds}ds\right)\left(\hat{\mathbf{T}}(s)+\frac{d\hat{\mathbf{T}}(s)}{ds}ds\right)$$
$$-\mu|\mathbf{N}_o|\left(\hat{\mathbf{T}}(s)+\frac{d\hat{\mathbf{T}}(s)}{ds}\cdot\frac{ds}{2}\right)=0$$
(A.2)

After eliminating the higher-order small terms, Eq. (A.2) takes following form,

$$-F(s)\hat{\mathbf{T}}(s)+|\mathbf{N}_o|\left(\hat{\mathbf{N}}(s)+\frac{d\hat{\mathbf{N}}(s)}{ds}\cdot\frac{ds}{2}\right)+F(s)\hat{\mathbf{T}}(s)+F(s)\frac{d\hat{\mathbf{T}}(s)}{ds}ds+\hat{\mathbf{T}}(s)\frac{dF(s)}{ds}ds$$
$$+\left(\frac{dF(s)}{ds}ds\right)\left(\frac{d\hat{\mathbf{T}}(s)}{ds}ds\right)-\mu|\mathbf{N}_o|\left(\hat{\mathbf{T}}(s)+\frac{d\hat{\mathbf{T}}(s)}{ds}\cdot\frac{ds}{2}\right)=0$$
(A.3)

$$F(s)\frac{d\hat{\mathbf{T}}(s)}{ds}ds+\hat{\mathbf{T}}(s)\frac{dF(s)}{ds}ds+|\mathbf{N}_o|\left(\hat{\mathbf{N}}(s)+\frac{d\hat{\mathbf{N}}(s)}{ds}\cdot\frac{ds}{2}\right)-\mu|\mathbf{N}_o|\left(\hat{\mathbf{T}}(s)+\frac{d\hat{\mathbf{T}}(s)}{ds}\cdot\frac{ds}{2}\right)=0$$
(A.4)

Frenet-Serret formula,

$$\begin{Bmatrix}\dot{\hat{\mathbf{T}}}(s)\\ \dot{\hat{\mathbf{N}}}(s)\\ \dot{\hat{\mathbf{B}}}(s)\end{Bmatrix}=\begin{bmatrix}0 & \kappa & 0\\ -\kappa & 0 & \tau\\ 0 & -\tau & 0\end{bmatrix}\begin{Bmatrix}\hat{\mathbf{T}}(s)\\ \hat{\mathbf{N}}(s)\\ \hat{\mathbf{B}}(s)\end{Bmatrix}$$
(A.5)

Substituting the above expressions in Eq. (A.4)

$$F(s)\kappa\hat{\mathbf{N}}(s)ds+\hat{\mathbf{T}}(s)\frac{dF(s)}{ds}ds+|\mathbf{N}_o|\left(\hat{\mathbf{N}}(s)+\left(-\kappa\hat{\mathbf{T}}(s)+\tau\hat{\mathbf{B}}(s)\right)\cdot\frac{ds}{2}\right)-\mu|\mathbf{N}_o|\left(\hat{\mathbf{T}}(s)+\kappa\hat{\mathbf{N}}(s)\cdot\frac{ds}{2}\right)=0$$
(A.6)

Dot multiplication of Eq. (A.6) with T(s) yields,

$$0+\frac{dF(s)}{ds}ds+|\mathbf{N}_o|\left(0+(-\kappa+0)\cdot\frac{ds}{2}\right)-\mu|\mathbf{N}_o|\left(1+0\cdot\frac{ds}{2}\right)=0$$
(A.7)

$$\frac{dF(s)}{ds}ds-|\mathbf{N}_o|\kappa\frac{ds}{2}-\mu|\mathbf{N}_o|=0$$
(A.8)

Dot multiplication of Eq. (A.6) with N(s) yields,

$$F(s)\kappa ds + |\mathbf{N}_o| - \mu |\mathbf{N}_o| \left( \kappa \frac{ds}{2} \right) = 0 \tag{A.9}$$

$$\mu \frac{dF(s)}{ds} ds - F(s)\kappa ds - |\mathbf{N}_o|(1+\mu^2) = 0 \tag{A.10}$$

$$|\mathbf{N}_o| = \frac{\mu}{(1+\mu^2)} \frac{dF(s)}{ds} ds - \frac{\kappa}{(1+\mu^2)} F(s) ds = \alpha \frac{dF(s)}{ds} ds - \beta F(s) ds \tag{A.11}$$

where $\alpha = \dfrac{\mu}{(1+\mu^2)}$ and $\beta = \dfrac{\kappa}{(1+\mu^2)}$

$$F(s)\kappa ds + \alpha \frac{dF(s)}{ds} ds - \beta F(s) ds - \mu\alpha \frac{dF(s)}{ds} ds.\kappa.\frac{ds}{2} + \mu\beta F(s) ds.k.\frac{ds}{2} = 0 \tag{A.12}$$

$$\alpha \frac{dF(s)}{ds} = (\beta - \kappa) F(s) \tag{A.13}$$

$$\int \frac{dF(s)}{F(s)} = \int \frac{(\beta - \kappa)}{\alpha} ds \tag{A.14}$$

$$\ln \frac{F_o}{F_I} = \int \frac{(\beta - \kappa)}{\alpha} ds \tag{A.15}$$

$$\frac{F_o}{F_I} = e^{\int \frac{-(\kappa - \beta)}{\alpha} ds} \tag{A.16}$$

$$\frac{F_o}{F_I} = e^{\int \frac{-\left(\kappa - \frac{\kappa}{1+\mu^2}\right)}{\frac{\mu}{1+\mu^2}} ds} = e^{\int -\mu\kappa ds} = e^{-\mu\phi_{bend}} \tag{A.17}$$

Eq. (A.17) can be used to determine the transmitted force at other end of the tendon-sheath actuator system.

## 2. Experimental characterization of tendon-sheath system

The friction coefficient between tendon and sheath is determined with the aid of an experimental setup as shown in Fig.13a and 13b. The bicycle brake wire is used as a transmission system, which is connected with the stepper motor at one end and weight at another end.

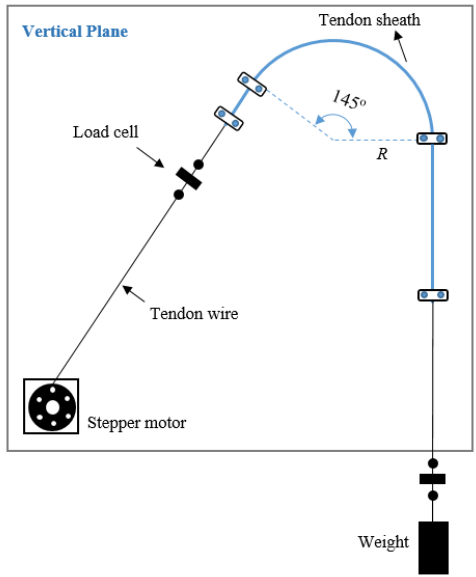
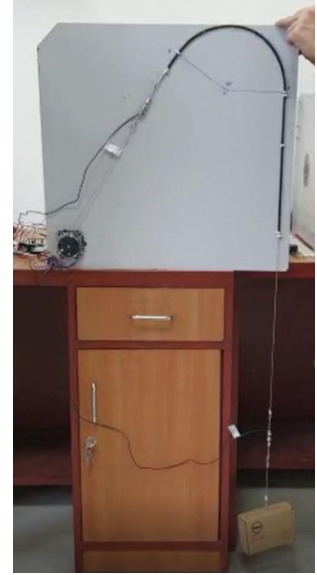

**Fig. 13a.** Schematic diagram of experimental setup

**Fig. 13b.** Actual experimental setup in laboratory

In this study, two scenarios are examined in which the motor is first rotated in the clockwise direction and then in the counterclockwise direction. The readings of two load-cells are presented in Table-A.1.

**Table A.1.** Readings of load-cells (kg) in two cases.

| Case 1* | | Case 2** | |
|---|---|---|---|
| Load cell –I (kg) | Load cell –II (kg) | Load cell –I (kg) | Load cell –II (kg) |
| 1.236 | 1.028 | 0.858 | 1.036 |
| 1.236 | 1.027 | 0.858 | 1.036 |
| 1.236 | 1.028 | 0.857 | 1.036 |

*weight is moving in upward direction
**weight is moving in downward direction

Case 1.
$F_o = F_I e^{-\mu\theta}$
$F_o = 1.028$
$F_I = 1.236$

Case 2.
$F_o = F_I e^{\mu\theta}$
$F_o = 1.036$
$F_I = 0.858$

$$e^{\mu\theta} = \frac{1.236}{1.028} = 1.202$$
$$\mu = 0.073$$

$$e^{\mu\theta} = \frac{1.036}{0.858} = 1.208$$
$$\mu = 0.075$$

The coefficient of friction ($\mu$) between brake wire and conduit is obtained as 0.073 and 0.075 in case 1 and 2, respectively.

**Data availability statement**

The data that support the findings of this study will be available upon request.

**Conflict of interest**

The authors declare that there is no conflict of interest regarding the publication of this paper.

**References:**

[1] Petersen KH, Shepherd RF. Fluid-driven intrinsically soft robots. Robot. Syst. Auton. Platforms, Elsevier; 2019, p. 61–84. https://doi.org/10.1016/B978-0-08-102260-3.00004-4.

[2] Davis S, Tsagarakis N, Canderle J, Caldwell DG. Enhanced modelling and performance in braided pneumatic muscle actuators. Int J Rob Res 2003;22:213–27. https://doi.org/10.1177/0278364903022003006.

[3] Kumar D, Ghosh S, Roy S, Santapuri S. Modeling and analysis of an electro-pneumatic braided muscle actuator. J Intell Mater Syst Struct 2021;32:399–409. https://doi.org/10.1177/1045389X20953624.

[4] Cafolla D, Russo M, Carbone G. CUBE, a cable-driven device for limb rehabilitation. J Bionic Eng 2019;16:492–502.

[5] Laribi MA, Carbone G, Zeghloul S. On the optimal design of cable driven parallel robot with a prescribed workspace for upper limb rehabilitation tasks. J Bionic Eng 2019;16:503–13.

[6] Phee SJ, Low SC, Dario P, Menciassi A. Tendon sheath analysis for estimation of distal end force and elongation for sensorless distal end. Robotica 2010;28:1073–82. https://doi.org/10.1017/S026357470999083X.

[7] Hellman RB, Santos VJ. Design of a back-driveable actuation system for modular control of tendon-driven robot hands. 2012 4th IEEE RAS EMBS Int. Conf. Biomed. Robot. Biomechatronics, IEEE; 2012, p. 1293–8. https://doi.org/10.1109/BioRob.2012.6290939.

[8] Chen L, Wang X. Modeling of the tendon-sheath actuation system. 2012 19th Int. Conf. Mechatronics Mach. Vis. Pract., IEEE; 2012, p. 489–94.

[9] Wang Z, Sun Z, Phee SJ. Modeling tendon-sheath mechanism with flexible configurations for robot control. Robotica 2013;31:1131–42.

https://doi.org/10.1017/S0263574713000386.

[10] Sun Z, Wang Z, Phee SJ. Elongation modeling and compensation for the flexible tendon-sheath system. IEEE/ASME Trans Mechatronics 2013;19:1243–50. https://doi.org/10.1109/TMECH.2013.2278613.

[11] Do TN, Tjahjowidodo T, Lau MWS, Phee SJ. A new approach of friction model for tendon-sheath actuated surgical systems: Nonlinear modelling and parameter identification. Mech Mach Theory 2015;85:14–24. https://doi.org/10.1016/j.mechmachtheory.2014.11.003.

[12] Lau KC, Leung EYY, Poon CCY, Chiu PWY, Lau JYW, Yam Y. Motion Compensation of Tendon-Sheath Driven Continuum Manipulator for Endoscopic Surgery. MATEC Web Conf., vol. 32, EDP Sciences; 2015, p. 4007. https://doi.org/10.1051/matecconf/20153204007.

[13] Xue R, Ren B, Yan Z, Du Z. A cable-pulley system modeling based position compensation control for a laparoscope surgical robot. Mech Mach Theory 2017;118:283–99. https://doi.org/10.1016/j.mechmachtheory.2017.08.006.

[14] Wagner CR, Emmanouil E. Efficiency and power limits of electrical and tendon-sheath transmissions for surgical robotics. Front Robot AI 2018;5:50. https://doi.org/10.3389/frobt.2018.00050.

[15] Shao Z, Wu Q, Chen B, Wu H, Zhang Y. Modeling and inverse control of a compliant single-tendon-sheath artificial tendon actuator with bending angle compensation. Mechatronics 2019;63:102262. https://doi.org/10.1016/j.mechatronics.2019.102262.

[16] Ismail R, Ariyanto M, Perkasa IA, Adirianto R, Putri FT, Glowacz A, et al. Soft elbow exoskeleton for upper limb assistance incorporating dual motor-tendon actuator. Electronics 2019;8:1184. https://doi.org/10.3390/electronics8101184.

[17] Jia X, Zhang Y, Jiang J, Du H, Yu Y. Design and analysis of a novel long-distance double tendon-sheath transmission device for breast intervention robots under MRI field. Adv Mech Eng 2020;12:1687814020904565. https://doi.org/10.1177/1687814020904565.

[18] Yin M, Wu H, Xu Z, Han W, Zhao Z. Compliant control of single tendon-sheath actuators applied to a robotic manipulator. IEEE Access 2020;8:37361–71. https://doi.org/10.1109/ACCESS.2020.2973173.

[19] Phan PT, Thai MT, Hoang TT, Lovell NH, Do TN. HFAM: Soft hydraulic filament artificial muscles for flexible robotic applications. Ieee Access 2020;8:226637–52. https://doi.org/10.1109/ACCESS.2020.3046163.

[20] Ennaiem F, Chaker A, Sandoval J, Bennour S, Mlika A, Romdhane L, et al. Cable-Driven parallel robot workspace identification and optimal design based on the upper limb functional rehabilitation. J Bionic Eng 2022;19:390–402.

[21] Lee D-H, Kim Y-H, Collins J, Kapoor A, Kwon D-S, Mansi T. Non-linear hysteresis compensation of a tendon-sheath-driven robotic manipulator using motor current. IEEE Robot Autom Lett 2021;6:1224–31. https://doi.org/10.1109/LRA.2021.3057043.

[22] Wu H, Yin M, Xu Z, Zhao Z, Han W. Transmission characteristics analysis and

compensation control of double tendon-sheath driven manipulator. Sensors 2020;20:1301. https://doi.org/10.3390/s20051301.

[23] Jung Y, Bae J. Torque control of a series elastic tendon-sheath actuation mechanism. IEEE/ASME Trans Mechatronics 2020;25:2915–26. https://doi.org/10.1109/TMECH.2020.2997945.

[24] Hoang TT, Sy L, Bussu M, Thai MT, Low H, Phan PT, et al. A wearable soft fabric sleeve for upper limb augmentation. Sensors 2021;21:7638. https://doi.org/10.3390/s21227638.

[25] Kernot JE, Ulrich S. Adaptive Control of a Tendon-Driven Manipulator for Capturing Non-Cooperative Space Targets. J Spacecr Rockets 2022;59:111–28. https://doi.org/10.2514/1.A34881.

[26] Bardi E, Gandolla M, Braghin F, Resta F, Pedrocchi ALG, Ambrosini E. Upper limb soft robotic wearable devices: a systematic review. J Neuroeng Rehabil 2022;19:1–17. https://doi.org/10.1186/s12984-022-01065-9.

[27] Zhang C. General compensation control method of flexible manipulator driven by tendon-sheath mechanism. J. Phys. Conf. Ser., vol. 2364, IOP Publishing; 2022, p. 12019. https://doi.org/10.1088/1742-6596/2364/1/012019.

[28] Han S, Wang H, Tian Y, Yu H. Enhanced extended state observer-based model-free force control for a Series Elastic Actuator. Mech Syst Signal Process 2023;183:109584. https://doi.org/10.1016/j.ymssp.2022.109584.

[29] Zhang Z, Zhang G, Wang S, Shi C. Hysteresis Modelling and Compensation for Tendon-Sheath Mechanisms in Robot-Assisted Endoscopic Surgery Based on the Modified Bouc-Wen Model with Decoupled Model Parameters. IEEE Trans Med Robot Bionics 2023. https://doi.org/10.1109/TMRB.2023.3249234.

[30] Yuan J, Fan Y, Wu Y. Design of motor cable artificial muscle (MC-AM) with tendon sheath–pulley system (TSPS) for musculoskeletal robot. Robotica 2023:1–17. https://doi.org/10.1017/S026357472300005X.

[31] Abdelhafiz MH, Andreasen Struijk LNS, Dosen S, Spaich EG. Biomimetic Tendon-Based Mechanism for Finger Flexion and Extension in a Soft Hand Exoskeleton: Design and Experimental Assessment. Sensors 2023;23:2272. https://doi.org/10.3390/s23042272.

[32] Ramasamy A, Design and development of an external wearable exo-suit for upper limb power augmentation. MTech Thesis 2021.